\documentclass[12pt]{article}
\usepackage{lmodern}
\usepackage{amssymb,amsmath}
\usepackage{ifxetex,ifluatex}
\usepackage{fixltx2e} 
\ifnum 0\ifxetex 1\fi\ifluatex 1\fi=0 
  \usepackage[T1]{fontenc}
  \usepackage[utf8]{inputenc}
\else 
  \ifxetex
    \usepackage{mathspec}
  \else
    \usepackage{fontspec}
  \fi
  \defaultfontfeatures{Ligatures=TeX,Scale=MatchLowercase}
\fi
\IfFileExists{upquote.sty}{\usepackage{upquote}}{}
\IfFileExists{microtype.sty}{%
\usepackage{microtype}
\UseMicrotypeSet[protrusion]{basicmath} 
}{}
\usepackage[margin=1in]{geometry}
\usepackage{hyperref}
\hypersetup{unicode=true,
            pdftitle={Deducing neighborhoods of classes from a fitted model},
            pdfauthor={Alexander Gerharz,\^{}\{a,*\}, Andreas Groll,\^{}a, Gunther Schauberger,\^{}b},
            pdfborder={0 0 0},
            breaklinks=true}
\usepackage{natbib}
\usepackage{graphicx,grffile}
\makeatletter
\def\maxwidth{\ifdim\Gin@nat@width>\linewidth\linewidth\else\Gin@nat@width\fi}
\def\maxheight{\ifdim\Gin@nat@height>\textheight\textheight\else\Gin@nat@height\fi}
\makeatother
\setkeys{Gin}{width=\maxwidth,height=\maxheight,keepaspectratio}
\IfFileExists{parskip.sty}{%
\usepackage{parskip}
}{
\setlength{\parindent}{0pt}
\setlength{\parskip}{6pt plus 2pt minus 1pt}
}
\ifx\paragraph\undefined\else
\let\oldparagraph\paragraph
\renewcommand{\paragraph}[1]{\oldparagraph{#1}\mbox{}}
\fi
\ifx\subparagraph\undefined\else
\let\oldsubparagraph\subparagraph
\renewcommand{\subparagraph}[1]{\oldsubparagraph{#1}\mbox{}}
\fi

\let\rmarkdownfootnote\footnote%
\def\footnote{\protect\rmarkdownfootnote}

\usepackage{titling}

  \title{Deducing neighborhoods of classes from a fitted model}
    \pretitle{\vspace{\droptitle}\centering\huge}
  \posttitle{\par}
    \author{Alexander Gerharz,\(^{a,*}\), Andreas Groll,\(^a\), Gunther
Schauberger,\(^b\)}
    \preauthor{\centering\large\emph}
  \postauthor{\par}
      \predate{\centering\large\emph}
  \postdate{\par}
    \date{11 September 2020}

\usepackage{amsmath}
\usepackage{graphicx}
\usepackage{csquotes}
\usepackage{subcaption}
\usepackage{fancyhdr}
\usepackage{color}

\DeclareMathOperator*{\argmax}{arg\,max}
\definecolor{grey}{rgb}{0.5,0.5,0.5}

\begin{document}
\maketitle

\footnotesize{a) TU Dortmund University, Germany; $^*$corresponding author: gerharz@statistik.tu-dortmund.de}

\footnotesize{b) Technical University of Munich, Germany}

\section*{Abstract}\label{abstract}

In todays world the request for very complex models for huge data sets
is rising steadily. The problem with these models is that by raising the
complexity of the models, it gets much harder to interpret them. The
growing field of \textit{interpretable machine learning} tries to make
up for the lack of interpretability in these complex (or even blackbox-)
models by using specific techniques that can help to understand those
models better. In this article a new kind of interpretable machine
learning method is presented, which can help to understand the
partitioning of the feature space into predicted classes in a
classification model using quantile shifts. To illustrate in which
situations this quantile shift method (QSM) could become beneficial, it
is applied to a theoretical medical example and a real data example.
Basically, real data points (or specific points of interest) are used
and the changes of the prediction after slightly raising or decreasing
specific features are observed. By comparing the predictions before and
after the manipulations, under certain conditions the observed changes
in the predictions can be interpreted as neighborhoods of the classes
with regard to the manipulated features. Chordgraphs are used to
visualize the observed changes.

\textbf{Keywords}: Interpretable Machine Learning, Explainable
Artificial Intelligence, Classification Task, Feature Space
Partitioning, Chordgraphs

\hypertarget{introduction}{%
\section{\texorpdfstring{Introduction\label{intro}}{Introduction}}\label{introduction}}

With the increasing demand for very complex models in the areas of data
analysis and predictive modeling the number of blackbox models is
growing steadily. The problem with these models is that by raising the
predictive power of a model or an algorithm by adding more complexity to
it, the loss of interpretability can be tremendous. Often it is easy to
understand the general idea of the fitting algorithm, but to understand
every single detail of the prediction process of the specific model is
pretty hard. In a random forest with 500 trees, for example, it is easy
to understand a single classification tree, but to completely understand
the whole ensemble model it is necessary to look at every split in every
tree, which gets too expensive if the corresponding classification task
was very huge and complex \citep{Breiman2001}.

\begin{figure}[h]
  \centering
    \includegraphics[width = 0.6\textwidth]{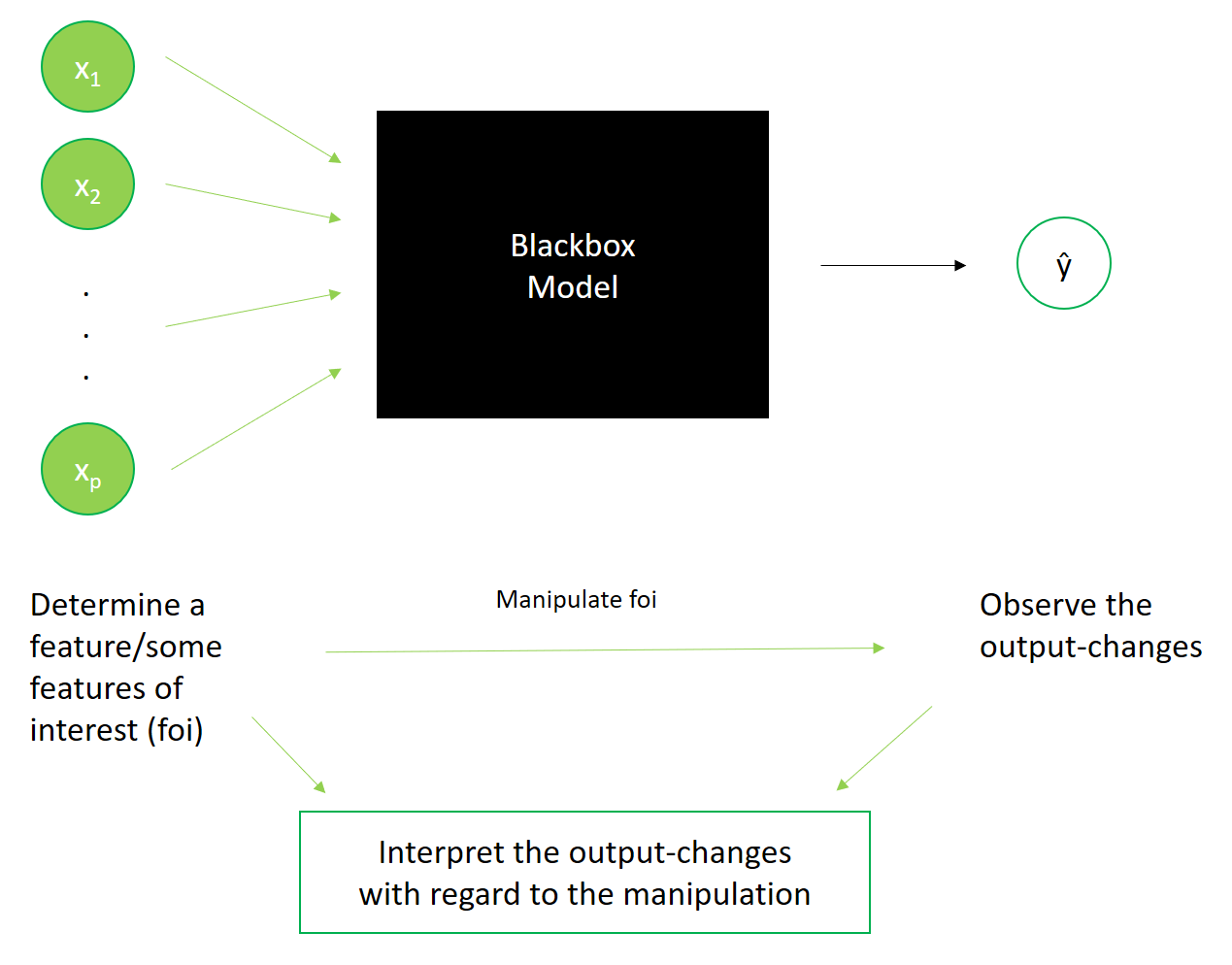}
  \caption{Basic concept behind most interpretable machine learning methods/explainable artificial intelligence}
  \label{iml_concept}
\end{figure}

The world of interpretable machine learning (IML) methods tries to open
a door to understand the internals of these complex models without
having to understand every single internal detail of them. A famous IML
method is the computation of the permutation feature importance as
described by \citet{Breiman2001}. Here, the input is randomly permuted
feature by feature and the increase of the misclassification rate is
measured to determine the features' importance in the model. In
contrast, the partial dependence plot, for example, does not calculate
the importance of a feature in a model, but it is a well-known method to
estimate the mean effect of a specific feature on the target value by
manipulating the inputs of some data and to observe how the output
changes \citep{Friedman2001}. A typical structure for these kind of IML
methods is displayed in Figure~\ref{iml_concept}.

Another interpretable machine learning method that is based on this
structure is the individual conditional expectations (ICE) plot, which,
similar to the partial dependence plot, describes the effect of a
specific feature on the target value, but instead of displaying a mean
effect it presents the individual changes for every observation
\citep{Goldstein2015}. Another completely different IML method is the
usage of anchors \citep{Ribeiro2018}. Anchors are used to find specific
features and their respective feature values that determine the
prediction of an observation, while the other features could be randomly
altered without affecting the prediction too much.

A first attempt to summarize the yet rather limited selection of
available IML methods is found in the publicly available book of
\citet{Molnar}, which lists more IML methods and explains their usages
on every day examples. Most of those methods are applicable on both
regression and classification tasks (or even more), while the method
proposed in this article is specifically designed for classification
tasks only.

The quantile shift method (QSM) presented in this work is based on the
basic concept of IML (see Figure~\ref{iml_concept}) and
is used to determine which classes are modeled as neighbors by a fitted
model with regard to specific features of interest. The QSM is used to
determine, which small changes in the features lead to a substantial
change in the predictions as the predicted class labels change. These
changes can then be interpreted as neighborhoods for the different
classes of an observation before and after the manipulation. In
contrast, the anchors method is used to find the features and their
respective values, which determine a specific prediction and interpret
them as substantial for a specific prediction. While both methods
observe whether slight changes in the features change a prediction, the
interpretation is substantially different.

The remainder of this article is structured as follows. In
Section~\ref{method}, we introduce the mathematical
details of the method and derive the corresponding change matrix, which
will later be presented as a chordgraph. Additionally, we illustrate the
method's relevance with an artificially created example with labels from
the field of medicine and also provide an in-depth discussion and
explanation of how to generally interpret the method's results. In
Section~\ref{iris_section}, the method's effectiveness
is illustrated with a real data example and different ways to use the
QSM are shown. Finally, Section~\ref{chap:conclusion}
concludes and discusses advantages and disadvantages of the proposed
method.

\hypertarget{methodology}{%
\section{\texorpdfstring{Methodology\label{method}}{Methodology}}\label{methodology}}

In this section, we first set the mathematical background for the QSM
and explain how to interpret it. As there are certain conditions, which
have to be kept in mind to assure a nice interpretation of the results,
we will then explain some possible pitfalls and how they can be solved
or estimated.

\hypertarget{mathematical-background}{%
\subsection{\texorpdfstring{Mathematical
background\label{mathematical}}{Mathematical background}}\label{mathematical-background}}

In the following, we will set the mathematical background for the QSM.
The aim is to slightly increase or decrease the value of the features of
interest and observe the changes in the predicted classes.\\
Suppose \(\hat{f}(\mathbf{x})\) is a final model fitted for a
classification task on a sample of size \(n\) with \(K\) different
classes, \(K\geq 2\), and \(L\) be the set of all the features used for
this classification with a specific set-size \(p = |L|\). Then,
\(\hat\pi_{\hat{f},k}(\mathbf{x})\) denotes the estimated probability by
the model \(\hat{f}(\cdot)\) for an observation \(\mathbf{x}\) to belong
to a specific class \(k\in\{1,\ldots,K\}\). Next, we determine
\begin{align*}
k^*_{\hat{f}} (\mathbf{x}) = \argmax_k {\hat\pi_{\hat{f},k}(\mathbf{x})}
\end{align*} such that \(k^*_{\hat{f}} (\mathbf{x})\) is the class with
the highest probability as estimated by the model \(\hat{f}(\cdot)\) for
the observation \(\mathbf{x}\) (from here on we will always talk about
the same fitted model, which is why we drop index \(\hat{f}\) in the
following for better readability).\\

Next, we choose a subset \(M \subseteq L\) containing the features of
interest. Mostly, the subset \(M\) has a size of \(|M| = 1\), i.e.~we
focus on a single specific feature. Let now \(\tilde{\mathbf{x}}_i\)
represent the feature-vector for observation \(i\), \(i = 1, ..., n\),
where those features from M each were manipulated componentwisely by a
small amount.

\begin{figure}[h]
  \centering
    \includegraphics[width = 0.6\textwidth]{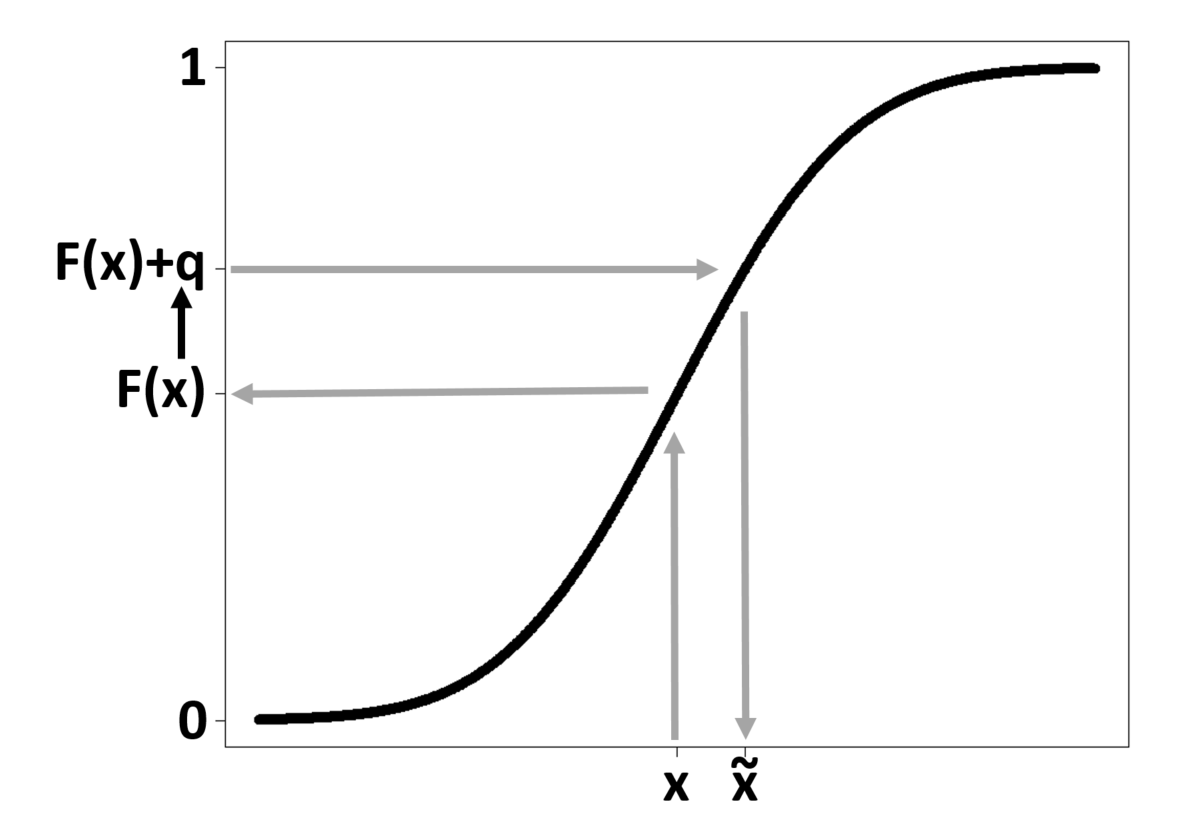}
  \caption{Example for a feature manipulation for a single feature (continuous case)}
  \label{distribution_x_q}
\end{figure}

The manipulation is done by slightly increasing or decreasing the
quantile-function of the subset \(M\) containing the features of
interest (see Figure~\ref{distribution_x_q}). For this
purpose, a small value \(q_l\), the quantile shift size, is added
componentwisely to \(\hat{F}_l(\cdot)\) denoting the empirical
cumulative distribution function (ecdf) for all features
\(l = 1, ..., p\), with \begin{align*}
q_l = \left\{
\begin{array}{ll}
u, & \text{\textit{for }}L_l \in M\text{\textit{, with }}u \in [-1,1]\\
0, & \, else. \\
\end{array}
\right.
\end{align*} To prevent extrapolation in the quantile function
\(\hat{F}^{-1}_l(\alpha)\), \(\alpha\) is chosen from the interval
\([0,1]\). Then, for a positive manipulation with \(q_l \in (0, 1]\), we
define: \begin{alignat*}{3}
 & \hat{F}_l(\tilde{x}_{i,l})    &&= min\{\hat{F}_l(x_{i,l}) + q_l, 1\} \nonumber \\
\Longrightarrow\quad & \tilde{x}_{i,l} &&= \hat{F}_l^{-1}(min\{\hat{F}_l(x_{i,l}) + q_l, 1\}).
\end{alignat*} The modifying values \(q_l\) for each \(l \in M\) are set
by the user. As this is a crucial point for the method, in the following
we provide some examples and recommendations for a reasonable choice of
\(q\).

The inverse of the ecdf \(\hat{F}_l^{-1}\) does not necessarily exist,
as \(\hat{F}_l\) typically is not continuous. Hence, for a positive
manipulation we have to define \begin{align}
\hat{F}_{l}^{-1}(\alpha) = inf\{x: \hat{F}_{l}(x) \ge \alpha\}.
\label{eq:quantile}
\end{align}

Equation~\eqref{eq:quantile} determines each value of
feature \(l\) after the manipulation as one out of the truly observed
values of the respective feature, which were used to estimate the ecdf.

Due to the definition of the inverse of the ecdf as defined in
Equation~\eqref{eq:quantile}, a positive manipulation is
generally not comparable to a negative manipulation, if it is done the
exact same way. While even a slight positive manipulation results in a
change of the corresponding feature's values, slight negative
manipulations typically change nothing at all. 

\begin{figure}[h]
  \centering
    \includegraphics[width = 0.8\textwidth]{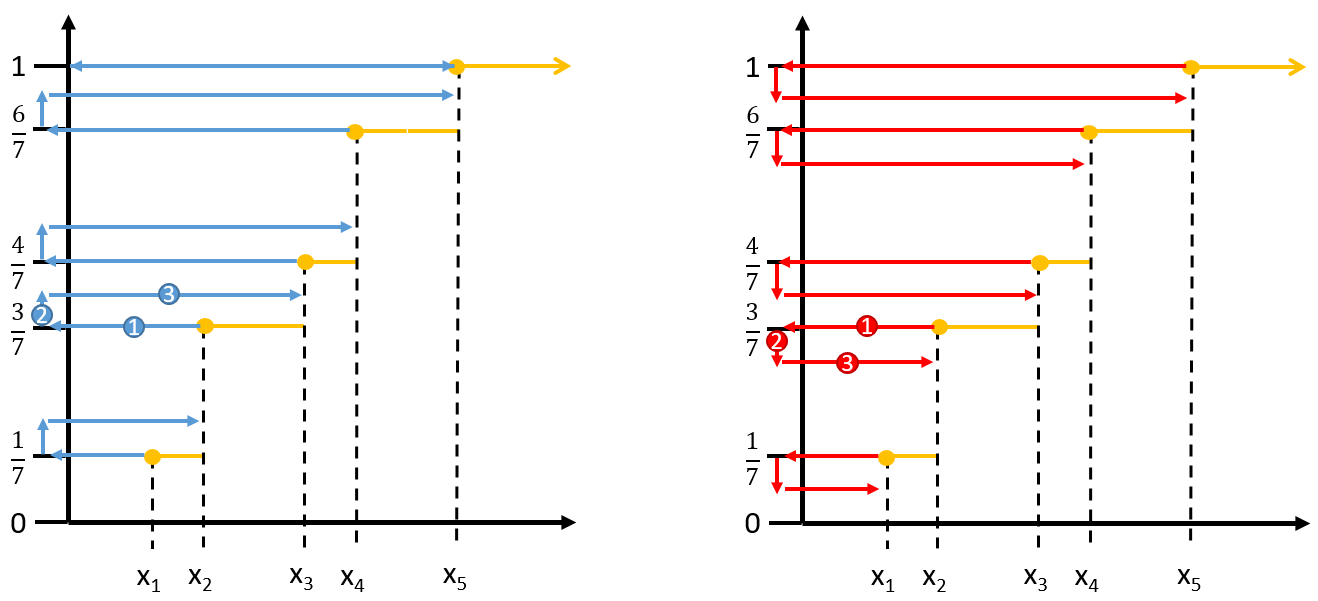}
  \caption{Comparison of positive (left) and negative (right) manipulation with the same $|q|$}
  \label{neg_mani}
\end{figure}

In Figure~\ref{neg_mani} a positive (left) and a negative (right) 
manipulation is shown for a specific example of a feature vector with 
five (unique) values, where the second and the fourth ordered value 
occur twice (see also Table~\ref{tab:negative_manip}). Now the QSM is used with $|q| < \frac{1}{7}$. In the ecdf as defined above, 
for the point $x_2$ for example, then $F(x_2) = \frac{3}{7}$ (blue arrow 1 in 
the left part of Figure~\ref{neg_mani}). If now the small 
$q$ is added this results in $\frac{3}{7} < \alpha < \frac{4}{7}$ (blue arrow 2). Due to the 
definition of Equation~\eqref{eq:quantile}, the positive manipulation results in 
$F^{-1}(\alpha) = x_3$ (blue arrow 3), which means that here even a  small positive manipulation 
results in a change of the feature value.

Next, assume that a small negative manipulation of equal size is
used, again for the feature value $x_2$ with corresponding ecdf of $F(x_2) = \frac{3}{7}$ (red arrow 1 in 
the right part of Figure~\ref{neg_mani}). 
Subtracting the amount $|q|$ now results in $\frac{2}{7} < \alpha < \frac{3}{7}$ (red arrow 2). 
However, due to the definition in Equation~\eqref{eq:quantile}, 
$F^{-1}(\alpha) = x_2$ (red arrow 3), which means the value has not changed at all. 
In fact, in this data example when subtracting $|q|$ not a single value would change and, hence,
negative and positive manipulations of the same absolute amount $|q|$ are typically not comparable.
For this reason, a negative manipulation has to be defined in another way.

A negative manipulation for a specific value \(x_{i,l}\) for observation
\(i\) and feature \(l\) can be done by using a positive manipulation for
all values of feature \(l\) and then searching for the minimal value
that is mapped to \(x_{i,l}\) (or, if no preimage for \(x_{i,l}\) is
found, the preimage of the next larger value with a corresponding
preimage is chosen). If this is done for all values of feature \(l\),
this is the corresponding negative manipulation, which is comparable to
a positive manipulation by the same absolute amount of \(q_l\).

\begin{table}[h]
\centering
\begin{tabular}{ c | c }
 Before & After \\ 
  \hline
 $x_1$ & $x_2$ \\  
 $x_2$ & $x_4$ \\  
 $x_2$ & $x_4$ \\  
 $x_3$ & $x_4$ \\  
 $x_4$ & $x_5$ \\  
 $x_4$ & $x_5$ \\  
 $x_5$ & $x_5$
\end{tabular}
\quad
\hspace{2cm}
\begin{tabular}{ c | c }
 Before & After \\ 
  \hline
 $x_1$ & $x_1$ \\  
 $x_2$ & $x_1$ \\  
 $x_2$ & $x_1$ \\  
 $x_3$ & $x_2$ \\  
 $x_4$ & $x_2$ \\  
 $x_4$ & $x_2$ \\  
 $x_5$ & $x_4$
\end{tabular}
\caption{Positive manipulation with $q = \frac{2}{7}$ (left); corresponding, intended negative manipulation with $q = -\frac{2}{7}$ (right)}\label{tab:negative_manip}
\end{table}

In Table~\ref{tab:negative_manip} a small example
illustrates this way of producing the negative manipulations. Without
loss of generality it is assumed that \(x_1 < x_2 < x_3 < x_4 < x_5\)
and the values \(x_2\) and \(x_4\) both exist twice in the dataset.
Next, the quantile shift size is chosen as \(q = -\frac{2}{7}\).
However, to obtain the resulting values of the negative manipulation, a
positive manipulation by the equivalent positive amount
\(|q| = \frac{2}{7}\) is done first (left part of
Table~\ref{tab:negative_manip}). Next, for every value
the preimage of the positive manipulation is searched for. If a value
does not have a preimage, then the next larger value with a
corresponding preimage is chosen and its lowest preimage is taken for
the manipulation. For the value \(x_4\), for example, the lowest value
from the set of corresponding preimages, i.e.~\(\{x_2,x_2,x_3\}\), which
is mapped to \(x_4\) by the positive manipulation is \(x_2\), which is
the corresponding value for the negative manipulation.

For \(q_l \in [-1, 0)\), this results in \begin{align*}
\tilde{x}_{i,l} = inf\{z \in \{x_{1,l}, ... x_{n,l}\}: \hat{F}_l^{-1}(min\{\hat{F}_l(z) + |q_l|, 1\}) \ge x_{i,l}\}.
\end{align*} Principally, it is recommended to choose
\(q = \pm \frac{v}{n}\) with \(v = 1, 2, ..., n\). Hence, \(v\) can be
chosen as the number of ordered values by which all the observations of
the respective feature of interest are shifted. When a tiny amount of
\(q\) is added to the ecdf, numerical machine arithmetic problems in the
sense of rounding
errors\footnote{The rounding errors are a consequence of the way real numbers are represented in a computer: as a signums-bit, a bit-sequence for the exponent and a bit-sequence for the significand.}
might occur when determining the manipulated feature value. Facing this
problem, the shift resulting from the addition might get larger than
intended.\}

To avoid this problem, the usage of \(q = \pm \frac{v}{n+1}\) is highly
recommended. Due to the definition in
Equation~\eqref{eq:quantile} this leads to the exact
same result as \(q = \pm \frac{v}{n}\), as \begin{align*}
\frac{v-1}{n} < \frac{v}{n+1} < \frac{v}{n},
\end{align*} for all \(v,n \in \mathbb{N}\) with \(v < n\).

Now, \(\tilde{\mathbf{x}}_{i}\) is the new manipulated observation,
which has the same value for those covariates from \(L\setminus M\) as
\(\mathbf{x}_{i}\), but different values for the features from \(M\).
These features were increased or decreased by the value that
corresponded to the componentwise raise or reduction of the respective
ecdf by the amount \(q_l\), not exceeding the minimum or maximum of the
empirical distribution of the features from \(M\).

Principally, this modus operandi does not only work for metric features,
but also for (ordered or nominal) categorical features of the form
\(x_l\in\{1,\ldots,c\}\), where \(c\) is the number of categories. For
this kind of features a manipulation from one group to another has to be
chosen manually, e.g.~switching from group \(r\) to another group \(s\)
within a specific feature \(l\), i.e.~changing from \(x_l=r\) to
\(x_l=s\).

Finally, for observation \(i, i = 1,...,n\), \(M \subseteq L\) and
\(\mathbf{q} = (q_1, ..., q_p)^T\) (possibly including some zeros if
\(M \subset L\)) let \(C_{\mathbf{q},M} (x_i)\) define the pair of the
original and the (potentially) new class prediction resulting from this
manipulation, i.e. \begin{align*}
C_{\mathbf{q},M} (\mathbf{x}_i)  &= \left( k^*(\mathbf{x}_i), k^* (\tilde{\mathbf{x}}_i) \right) \nonumber \\
                  &= \left( k^* (\mathbf{x}_{i,L\setminus M},\mathbf{x}_{i,M}) , k^* (\tilde{\mathbf{x}}_{i,L\setminus M},\tilde{\mathbf{x}}_{i,M}) \right) \nonumber \\
                  &= \left( k^* (\mathbf{x}_{i,L\setminus M},\mathbf{x}_{i,M}), k^* (\mathbf{x}_{i,L\setminus M},\tilde{\mathbf{x}}_{i,M}) \right) \nonumber \\
                  &= (\hat{y}_{i,old},\hat{y}_{i,new}).
\end{align*} We obtain \(\hat{y}_{i,old} = \hat{y}_{i,new}\), if the
predicted class \textbf{has not changed} by manipulating
\(\mathbf{x}_{i,M}\) and \(\hat{y}_{i,old} \neq \hat{y}_{i,new}\)
otherwise. Note that
\(\tilde{\mathbf{x}}_{i,L\setminus M} = \mathbf{x}_{i,L\setminus M}\)
holds, as the features from \(L \setminus M\) were not modified.

\hypertarget{interpretation}{%
\subsection{\texorpdfstring{Interpretation\label{method_interpretation}}{Interpretation}}\label{interpretation}}

The results could now be given in form of a migration matrix for all
observations \(i = 1, ..., n\), where the rows indicate the predicted
classes of an observation before the manipulation of
\(\mathbf{x}_{i,M}\) and the columns indicate its predicted classes
after the manipulation. The trace of this migration matrix counts the
number of observations that have not changed classes by the
manipulation, while off-diagonal elements aggregate the number of
observations that have changed from the predicted class as indicated by
the respective row into the predicted class as indicated by the
respective column. An example of a migration matrix can be found in
Table~\ref{tab:migration}.

\begin{table}[h]
\centering
\begin{tabular}{ c c c }
  & $A_{after}$ & $B_{after}$ \\ 
  \hline
 $A_{before}$ & $n_{A \rightarrow A}$ & $n_{A \rightarrow B}$ \\  
 $B_{before}$ & $n_{B \rightarrow A}$ & $n_{B \rightarrow B}$    
\end{tabular}
\caption{General structure of migration matrices for 2 classes}\label{tab:migration}
\end{table}

The off-diagonal elements of Table~\ref{tab:migration}
can be interpreted as follows:

\begin{itemize}
  \item if $n_{A \rightarrow B} > 0$, there exists an area in which class $B$ is classified close to an area in which class $A$ is classified with regard to the manipulation of the features from $M$
  \item if $n_{A \rightarrow B} = 0$,  no area in which class $B$ is classified is found next to an area in which class $A$ is classified with regard to the manipulation of the features from $M$ - but it could still exist! (maybe the manipulation was not 
substantial enough and the data points have not reached the other side of the border 
or class $B$ was skipped because the manipulation was too strong)
\end{itemize}

Of course, \(n_{B \rightarrow A}\) can be interpreted analogously.

Chordgraphs are a nice way to represent these migration matrices. If we
have an examplary migration matrix as defined in
Table~\ref{tab:migration}, for a two class example the
migration matrix might look like
Table~\ref{tab:migrationexample}.

\begin{table}[h]
\centering
\begin{tabular}{ c c c }
  & $A_{after}$ & $B_{after}$ \\ 
  \hline
 $A_{before}$ & 10 & 0 \\  
 $B_{before}$ & 1 & 9    
\end{tabular}
\caption{Exemplary migration matrix for 2 classes}\label{tab:migrationexample}
\end{table}

\begin{figure}[h]
  \centering
    \includegraphics[width = 0.8\textwidth]{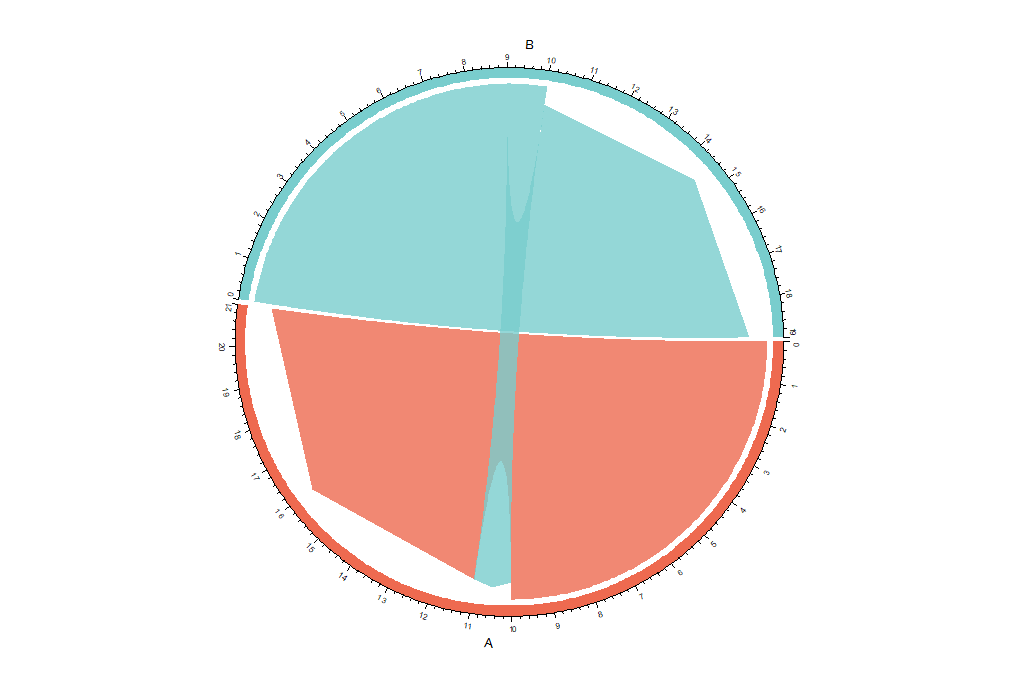}
  \caption{Chordgraph for migration matrix in Table~\ref{tab:migrationexample}}
  \label{chord_migrationexample}
\end{figure}

The migration matrix in Table~\ref{tab:migrationexample}
can be visualized as a chordgraph as shown in
Figure~\ref{chord_migrationexample}. In the lower half
of this figure the ten observations, which belong to class A before the
manipulation, are shown by the big red strang of chords starting on
the scale between 0 and 10 and ending up on the same scale between 11
and 21 as indicated by the arrow. This shows that ten observations belong to class A before
the manipulation and all ten observations are still in class A after the
manipulation. In the upper half of this figure the observations, which
belong to class B before the manipulation, are shown by the big
turquoise strang of chords starting on the scale between 0 and 10. From
this strang of chords a big part ends up on the upper halfs scale
between 10 and 19, which indicates that nine observations that belong to
class B before the manipulation are still in class B after the
manipulation, but a small part ends up in the lower half's scale between
10 and 11, which indicates that one observation belongs to class B
before the manipulation, but is in class A after the manipulation. This
is exactly, what the migration matrix in
Table~\ref{tab:migrationexample} indicates.

This shows, that the migration matrix, which was a result of the QSM
with a specific data manipulation, indicates that in the direction of
the data manipulation there is an area of class A modeled in the
direction of the manipulation next to an area of class B.

All plots and analyses in this work have been performed in R Version
3.6.0 \citep{R}. The chordgraphs, which are the main visualization tool
for this method, were computed with the \texttt{circlize}-package in R
\citep{circlize}.

\hypertarget{artifically-created-medical-application}{%
\subsection{\texorpdfstring{Artifically created medical
application\label{med_ex}}{Artifically created medical application}}\label{artifically-created-medical-application}}

First, we create a simple artificial data example and assign specific
labels to the classes in order to illustrate for which kind of research
questions this method is applicable. In the same context, we will show
why choosing quantiles as basis of the data manipulation can be
beneficial compared to choosing a specific number. For illustration
purposes, the present example is rather simple and clearly structured,
but especially in really complex data situations, in which there are too
many features to look at in single plots, using quantile-based
manipulations can be advantageous.

\begin{figure}[h]
  \centering
    \includegraphics[width = 0.6\textwidth]{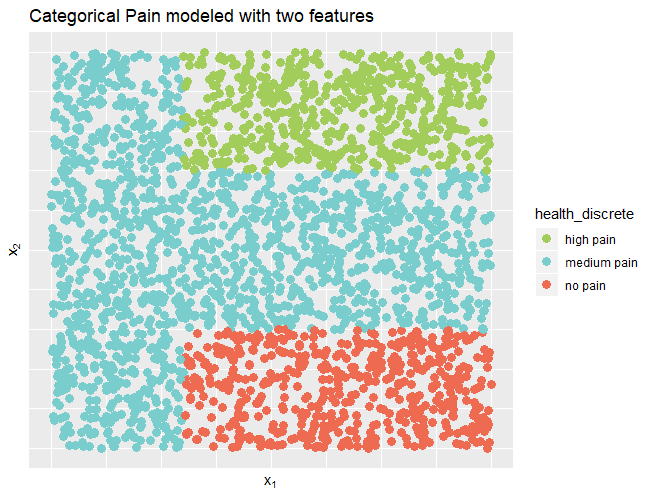}
  \caption{Predicted pain levels by two metrical features}
  \label{medical_example}
\end{figure}

Figure~\ref{medical_example} shows a 2-dimensional
feature space in which three different pain levels are predicted by some
statistical model for which we assume that it is able to describe the
relationship between \(x_1\), \(x_2\) and the target \(y\) very well.
The model maps patients with features \(x_1\) and \(x_2\) to a 3-class
target \(y\). All patients with a low value of \(x_1\) are assigned to
the class \textit{medium pain}. However, if \(x_1\) exceeds a certain
threshhold, patients with a very low \(x_2\) are assigned to class
\textit{no pain}, patients with a medium value of \(x_2\) are assigned
to class \textit{medium pain} and patients with a large value of \(x_2\)
are assigned to class \textit{high pain}.

Suppose that this model is provided to a physician, who plans to raise
\(x_1\) to ease the pain of a patient (e.g., if \(x_1\) is the
heart-rate, the physician might advise the patient to become physically
more active). If at the same time also \(x_2\) is small, the model
agrees with the physician's assumption and the patient in fact might get
better. However, in contrast to linear relationships as modeled e.g.~by
standard linear regression, increasing \(x_1\) would not always lead to
improvement here. For a patient with a rather large value of \(x_2\), an
increase of \(x_1\) might even result in a migration from the
\textit{medium pain} level to \textit{high pain}. In this case, a
detailed analysis of the migration matrix for increasing \(x_1\) would
reveal the more complex, non-linear relationships (see
Table~\ref{mig_mat_med_1}).

\begin{figure}[h!]
\begin{subfigure}{.5\textwidth}
  \centering
  \includegraphics[width = \textwidth]{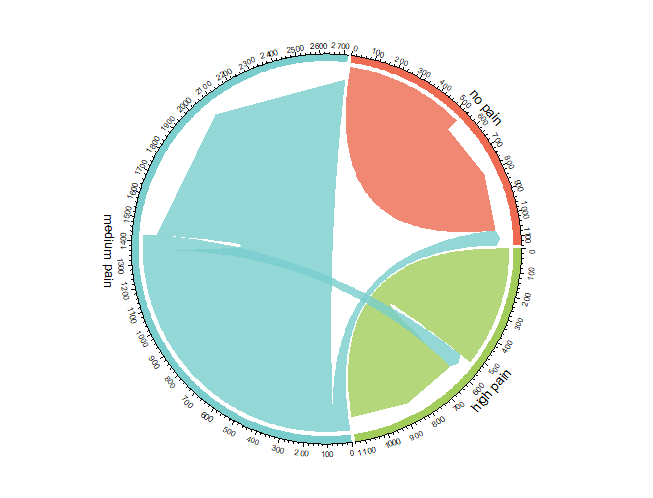}
  \caption{Changes in prediction when raising $x_1$ with $q_1 = \frac{251}{2500}$ and $x_2$ by $q_2 = 0$}
  \label{medical_chord}
\end{subfigure}
\hspace*{3mm}
\begin{subfigure}{.5\textwidth}
  \centering
  \includegraphics[width = \textwidth]{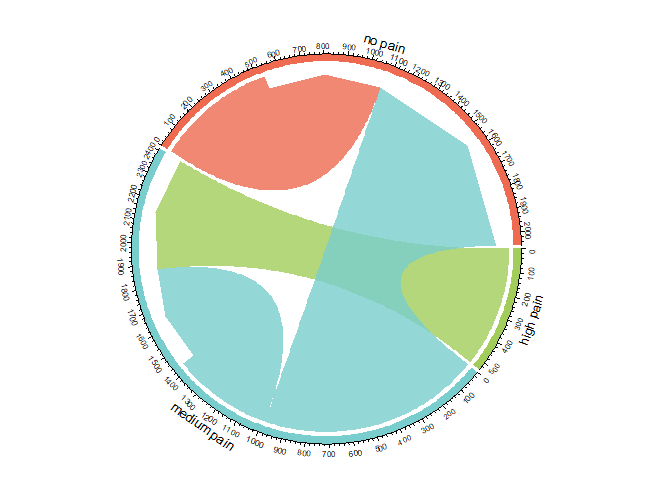}
  \caption{Changes in prediction when raising $x_1$ and decreasing $x_2$ with $q_1 = \frac{751}{2500}$ and $q_2 = -\frac{751}{2500}$}
  \label{medical_chord_big_manip}
\end{subfigure}
\caption{Artifically created medical data predicted with different data manipulations}
\label{medical_chord_fig}
\end{figure}

This migration-matrix is presented as a chordgraph in
Figure~\ref{medical_chord}, which is built up from all
the observations in the migration-matrix. Each observation is presented
as a single chord, starting from the class in which it was predicted
before and ending up in the class it was predicted after the
feature-manipulation. Multiple observations that have the same starting
and ending point in the chordgraph are combined together as a big strang
of chords, which is the wider the more observations have the same
starting and ending points.

In the present example, it is easy to see that after raising \(x_1\) by
just a small amount some \textit{medium pain} patients get better and
move to the \textit{no pain} category, while some get worse and end up
in the \textit{high pain} category.

Furthermore, it is displayed that by decreasing \(x_2\) only, a patient
with a \textit{high pain} or \textit{medium pain} level could actually
get better and by decreasing \(x_1\) only, \textit{high level} patients
could get better (see Figure~\ref{medical_example}).
Furthermore, by decreasing \(x_2\) and simultaneously increasing \(x_1\)
more substantially (e.g., in case the physician has a very effective
drug or any other method for the considerable manipulation of these
features) almost every patient classified in the \textit{high pain}
level class by the model is then assigned to a \textit{medium pain}
level and more than half of the \textit{medium pain} level predictions
get now predicted in the \textit{no pain} level category (see
Table~\ref{mig_mat_med_2} and
Figure~\ref{medical_chord_big_manip}).

In this case, if one would increase \(x_1\) and decrease \(x_2\) by a
small amount, then one can easily imagine that in the upper third of
Figure~\ref{medical_example} some of the medium pain
level patients could get worse and become high pain level patients. For
the chordgraph in Figure~\ref{medical_chord_big_manip}
the corresponding value of \(q\) was chosen so large that the graph
would not show this type of transition. Actually, the used manipulations
were too large such that the class of high pain level patients is
skipped. As mentioned in the remarks below, this method does not proof
that there is no neighborhood between the two classes for the regarded
manipulation. The chordgraph simply displays to which classes the
patients would migrate, when the manipulation of the features was that
large.

\begin{table}[h!]
\centering
\begin{tabular}{l|r r r}
\hline
  & high pain & medium pain & no pain\\
\hline
high pain & 534 & 0 & 0\\
medium pain & 73 & 1286 & 72\\
no pain & 0 & 0 & 535\\
\hline
\end{tabular}
\caption{Migration matrix for prediction changes when raising $x_1$ by $q_1=\frac{251}{2500}$ (and leaving $x_2$ unchanged).}
\label{mig_mat_med_1}
\end{table}
\vspace*{0.6cm}

\begin{table}[h!]
\centering
\begin{tabular}{l|r r r}
\hline
  & high pain & medium pain & no pain\\
\hline
high pain & 0 & 534 & 0\\
medium pain & 0 & 444 & 987\\
no pain & 0 & 0 & 535\\
\hline
\end{tabular}
\caption{Migration matrix for prediction changes when raising $x_1$ and at the same time decreasing $x_2$ by $q_1=\frac{751}{2500}$ and $q_2=-\frac{751}{2500}$, respectively.}
\label{mig_mat_med_2}
\end{table}
\bigskip

Note that in Figure~\ref{medical_example}, features
\(x_1\) and \(x_2\) were intentionally presented without a scale. The
intention is to illustrate by this artificial data example why using
quantile-based data manipulations can be advantegous compared to using
plain (rather arbitrary) numbers. The main reason is that it is often
hard to determine what size a \enquote{slight} increase or decrease
might be in the typical case when the exact distribution of the features
is unknown. This is particularly relevant when the task is to find
direct neighborhoods. Moreover, if many predictors are present, a fast
and automatic method for the computation of the corresponding
chordgraphs is essential, instead of determining for every feature and
observation manually, what a slight manipulation might be. Different
features usually have different scales and, depending on their location
in the feature space, a \enquote{slight} manipulation could have a
different meaning for different observations, especially if a feature of
interest has a very complex distribution (e.g., a multimodal
distribution). This can be achieved by using small amounts on the
quantile scale, which are comparable for all metric features.

Another situation when quantile-based modifications are preferable is
the presence of skewed features. A standard example in this regard could
be \textit{income}. Typically, the majority of individuals has a low to
moderate salary, whereas a few individuals have a (very) high income. So
while the change of the (e.g., monthly) income by a few hundred units
might have a drastic effect on the prediction of a certain factor
outcome for the first group of individuals, it might be too small to be
relevant for the prediction of the outcome of the high-earners. In
contrast, quantile-based modifications are more comparable for both
subintervals of the income distributions.

\hypertarget{remarks-about-the-methods-interpretation}{%
\subsection{Remarks about the method's
interpretation}\label{remarks-about-the-methods-interpretation}}

In the following, we give some important remarks regarding the
interpretation of the results.

\begin{enumerate}
\item If we define the preference order
\begin{align*}
A \succ B := \parbox{22em}{class $A$ is directly (or generally) next to class $B$ in
the direction of the manipulation,}
\end{align*}
then due to the fact that particularly complex models can produce also complex partitionings of the feature space, it follows
\begin{align*}
A \succ B \land B \succ C \not\Rightarrow A \succ C.
\end{align*}
This expresses that the results of the QSM can not be interpreted transitively. In particular, a rather complex model could classify a specific class as a small 
insula within another class or even spotwise in the feature space, in which case one could 
get results that seem transitive, but in fact are not (for more details, see Section~\ref{transitive}).

\item As indicated, the QSM is built to find neighborhoods 
as described by the model, but not to proof that there is no neighborhood 
between two classes regarding the manipulation of $\mathbf{x}_M$. 
If the goal is to \textbf{proof} that a certain class has no direct 
neighborhoods within the fitted model, 
then one would have to fill the complete modeled space of the 
class of interest with data points and then had to manipulate 
$\mathbf{x}_M$ with infinitely small steps from the starting points either until 
$+\infty$ or $-\infty$, respectively, in direction of the manipulation, or until all points have switched classes.

\item The manipulation of 
$\mathbf{x}_M$ could be too big, such that an intermediate class was skipped, 
and consequently, no direct neighborhood was found.
Hence, the \enquote{neighborhoods} from above should be regarded more generally as an \enquote{exists above} (if the manipulation was done by raising $\mathbf{x}_M$) or as an \enquote{exists below} (if the manipulation was done by decreasing $\mathbf{x}_M$). To find \textbf{direct} neighborhoods one would have to start with a very small manipulation of $\mathbf{x}_M$ and raise (or decrease) it continuously. In contrast to this, if one has a specific manipulation in mind, one could just use this specific manipulation and then the resulting migration matrix shows the corresponding 
class changes (if any).

\item If specific features are manipulated and a neighborhood is found between 
two classes, this neighborhood can indeed be interpreted as such, if the task at hand 
is to find out, how the  \textbf{model describes} the neighborhoods. But if the task at 
hand is to find realistic and practical neighborhoods between modeled classes, 
then these neighborhoods should always be investigated in two ways. If a neighborhood 
is found by the intended manipulation between two classes, this means that there 
exist data points on one side near the border between these two classes. This does 
not necessarily mean that on the other side of this border data points can also exist.
If similar manipulations are carried out in the opposite direction and this neighborhood 
is not confirmed, then this might mean that due to the manipulation unobservable 
feature combinations have been created and thus the found neighborhood has no 
practical use. 
A reason for this might be that the model 
simply extrapolates into this area 
of the feature space (general problem of 
extrapolation, which can lead to unreasonable interpretations; 
\citealp{Hooker}).

\item In many cases multiple data points could occur with equal values of a 
possible feature of interest. If those points are directly at a border between two classes, different 
problems can be observed, as shown in Section~\ref{multiple_datapoints}.

\item Finally, a rather straightforward and fundamental remark: 
If the model at hand is rather bad and inappropriate, the 
found neighborhoods between the classes are correct for describing 
and understanding the model, but would not reflect the reality. 
Hence, it is important to properly evaluate the model first, before it is interpreted!
\end{enumerate}

\hypertarget{transitive-interpretations}{%
\subsection{\texorpdfstring{Transitive
interpretations\label{transitive}}{Transitive interpretations}}\label{transitive-interpretations}}

\begin{figure}[h]
  \centering
    \includegraphics[width = 0.6\textwidth]{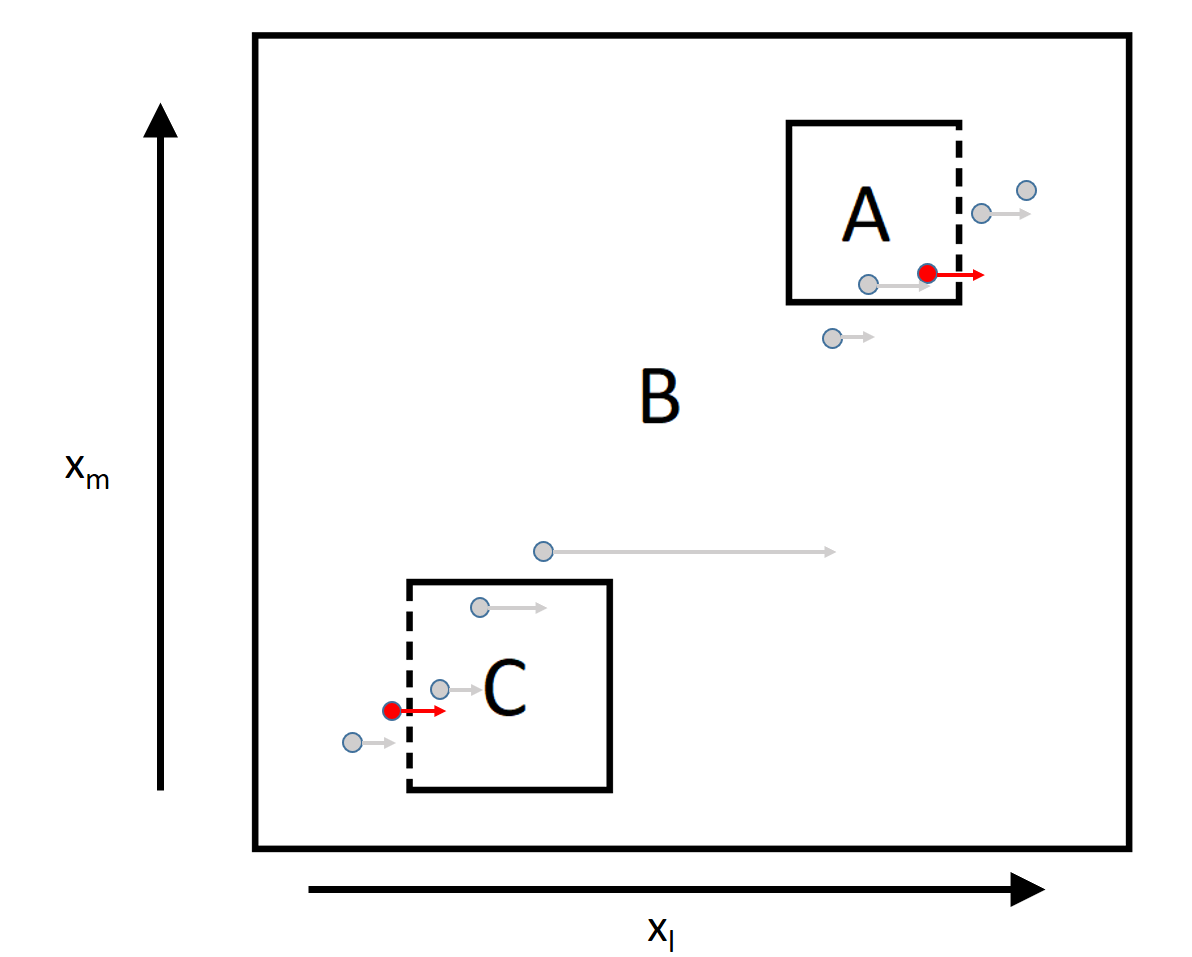}
  \caption{Example for transitivity problem in a 2-dimensional feature space}
  \label{transitivedots}
\end{figure}

To show possible problems regarding a transitive interpretation of this
method, here is a small artifical data example. In
Figure~\ref{transitivedots} a feature space with two
features \(x_{l}\) and \(x_{m}\) is shown. A model now labels most of
this feature space as class \textbf{B}, while a small area with a low
value of both features is labeled as class \textbf{C} and a small area
with high values of both features is labeled as class \textbf{A}. In
addition to that there are 10 red and grey data points, which are used
to describe the partitioning of the feature space with the QSM.

Now, the feature space partitioning should be determined by choosing
\(q_{x_{l}} = \frac{1}{11}\) and \(q_{x_{m}} = 0\), meaning keeping
\(x_{m}\) constant. Hence, only neighborhoods with regard to \(x_{l}\)
are looked for. With 10 different data points, this means that holding
\(x_{m}\) constant each data point gets assigned the next higher value
of \(x_{l}\) contained in the dataset. The data point with the highest
\(x_{l}\) does not change, as it is already at the upper bound of the
range of \(x_{l}\). This manipulation results in the following migration
matrix:

\begin{table}[h]
\centering
\begin{tabular}{ c c c c }
  & $A_{after}$ & $B_{after}$ & $C_{after}$ \\ 
  \hline
 $A_{before}$ & 1 & 1 & 0\\  
 $B_{before}$ & 0 & 5 & 1 \\   
 $C_{before}$ & 0 & 0 & 2
\end{tabular}
\caption{Migration matrix for QSM with the manipulation of $q = (\frac{1}{11}, 0)$ from Figure~\ref{transitivedots}}\label{tab:migrationtrans}
\end{table}

In Figure~\ref{transitivedots} the two points, which
change their prediction, are marked in red and switch the modeled class
through the dashed lines. These are the two points shown in the
migration matrix in Table~\ref{tab:migrationtrans} as
one has switched from class \textbf{A} to class \textbf{B} and the other
one from class \textbf{B} to class \textbf{C}. As there is an area of
class \textbf{B} modeled in the direction of the manipulation above an
area of class \textbf{A}, and then there is an area of class \textbf{C}
modeled in the direction of the manipulation above an area of class
\textbf{B}, based on the corresponding migration matrix one could
conclude that in the direction of the data manipulation there is some
kind of \enquote{hierarchy}. Particularly, here one might conclude that
along the direction of this manipulation class \textbf{A} is below class
\textbf{B}, which itself is below class \textbf{C}, but as shown in
Figure~\ref{transitivedots} this is not the case. Even
if the dimensionality would be too large to graphically visualize it,
just by checking the respective feature of interest for the groups
seperately would likely confirm the non-transitivity for this example.

\begin{figure}[h]
  \centering
    \includegraphics[width = 0.6\textwidth]{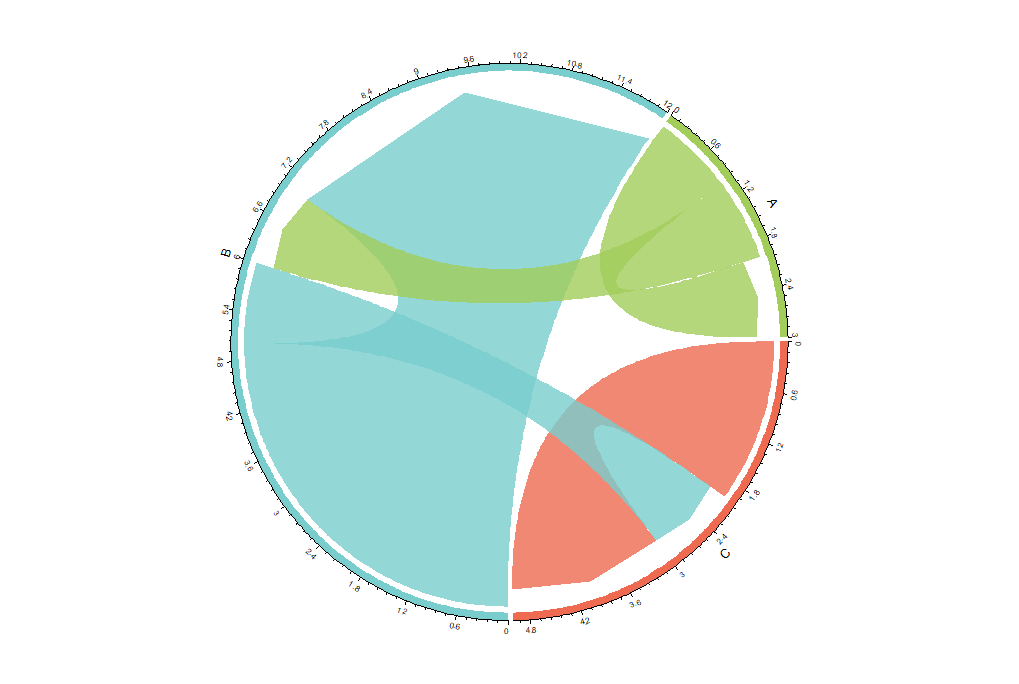}
  \caption{Chordgraph for migration matrix in Table~\ref{tab:migrationtrans}}
  \label{chord_transitive}
\end{figure}

In particular, the respective chordgraph as shown in
Figure~\ref{chord_transitive} can suggest some kind of
\enquote{hierarchy}. The chordgraph shows the migration of one
observation from class \textbf{A} to class \textbf{B} and the migration
of one observation from class \textbf{B} to class \textbf{C}, which
looks like a hierarchical structure that actually does not exist. To
conclude, QSM results should not be interpreted with regard to
transitivity!

In this specific case the QSM just describes that there is in fact an
area of class \textbf{B} modeled in the direction of the manipulation
above an area of class \textbf{A} and there is an area of class
\textbf{C} modeled in the direction of the manipulation above an area of
class \textbf{B}. Thus, with this specific manipulation these two
neighborhoods were found. When using other manipulations in other
directions, then other neighborhoods could be found.

\hypertarget{multiple-data-points-with-the-same-value}{%
\subsection{\texorpdfstring{Multiple data points with the same
value\label{multiple_datapoints}}{Multiple data points with the same value}}\label{multiple-data-points-with-the-same-value}}

\begin{figure}[h]
  \centering
    \includegraphics[width = 0.6\textwidth]{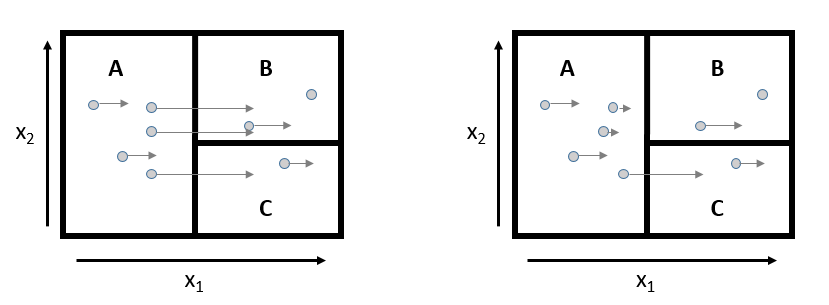}
  \caption{QSM with $q_{x_1} = \frac{1}{n+1}$ with some equal (left) and slightly jittered (right) data points}
  \label{multiple_at_once}
\end{figure}

When multiple data points with the same value occur in a dataset and the
QSM is used, some unexpected problems can occur. In the left graph of
Figure~\ref{multiple_at_once}, there are 3 data points
with an equal value for \(x_1\). When choosing the shift size
\(q_{x_1} = \frac{1}{n+1}\), so the smallest possible number that allows
the data points to shift, then all 3 data points would change their
predictions. In this case a neighborhood between class \textbf{A} and
class \textbf{B} and another neighborhood between class \textbf{A} and
class \textbf{C} would be found. Here, the corresponding migration
matrix and chordgraph would indicate a \enquote{stronger} neighborhood
between \textbf{A} and \textbf{B}, because more points switch between
these two classes than between class \textbf{A} and class \textbf{C}.
However, compared to the remaining unique data points the jumpsizes
(lengths of the arrows) of the three tied points are disproportionately
large. As shown in the right graph of
Figure~\ref{multiple_at_once}, if the same three data
points would not be exactly equal but just slightly differ, then just
one of these data points would change its prediction and the
neighborhood between class \textbf{A} and class \textbf{B} would not
even be found in this case, which is another problem.

If the shift size \(q_{x_1} = \frac{2}{n+1}\) is chosen, such that every
data point is shifted by 2 unique values of \(x_1\), then both
neighborhoods would be found even in the situation of the right graph in
Figure~\ref{multiple_at_once}, when the data points are
not exactly equal, as one of the data points changes its prediction from
class \textbf{A} to class \textbf{B}.

For any continuous (and random) feature the probability for a specific
value is zero. Hence, multiple data points with the same covariate value
theoretically should never occur. But in real world applications, for
example due to rounding, equal covariate values are possible and could
substantially change the interpretation of the QSM (see
Figure~\ref{multiple_at_once}).

If some observations have exactly the same value for a feature of
interest (e.g., due to rounding), then they tend to be shifted unfairly
compared to those observations with unique values for that feature. To
avoid this problem, there are some ways to adjust the QSM to treat
observations with equal values more fairly compared to those, which have
unique values for a specific feature.

\begin{enumerate}

  \item \underline{Shift all ties}: One possibility is to shift all of the 
  data points that share a specific value of a feature of interest and observe 
  the changes in the predictions (left graph in Figure~\ref{multiple_at_once}). 
  In this example this guarantees that all data points are shifted, and neighborhoods can be found more easily. As mentioned above, this method might tend to shift the groups of observations with equal values of a specific feature of interest 
  disproportionately large compared to observations with unique values.

  \item \underline{Repeatedly shift ties randomly}: Another alternative in the case of multiple observations with equal values for a specific feature of interest, is to repeatedly shift the 
observations. As shown in the right graph of Figure~\ref{multiple_at_once}, when the observations were just slightly jittered, in this example two observations were changed by 
  almost no amount and one by a larger amount. As a consqeuence, these observations are 
  treated more fair compared to the other observations than in the \textit{shift all ties} 
  method, but very unequally amongst themselves. For the artificial example, 
  the prediction of just one of these three obervations changes. 
  The \textit{repeatedly shift ties randomly} method does exactly that, but 
  instead of jittering and thus adding some blurredness to the data, it 
  repeatedly executes the QSM and randomly determines an order of the tied 
  observations. While all observations with unique values of the feature 
  of interest are shifted in the same way for all the repeated shifts, the 
  observations sharing their value of the feature of interest with other 
  observations are shifted in just a fraction of the repeats by the full shift as 
  in the \textit{shift all ties} method.
  Hence, observations with equal values of the specific feature of interest 
  are in most repetitions shifted less, and if $|q|$ is rather small, they might not be shifted at all. 
  Thus, in comparison to the \textit{shift all ties} 
  method, this method treats the observations with the same values of the 
  specific feature of interest more fair compared to the observations, 
  which have unique values of this feature of interest.

\end{enumerate}

The differences of these methods are illustrated in
Section~\ref{iris_section} on a real data example.

\hypertarget{real-data-application}{%
\section{\texorpdfstring{Real Data
Application\label{iris_section}}{Real Data Application}}\label{real-data-application}}

In this section, the QSM is applied to a real data set in order to
provide some insight on the method's potential. For this purpose, the
iris dataset is used, where the species of flowers are classified by
some metric features of the flowers.

The iris dataset is a dataset containing 150 observations for 3
different species of different iris flowers and 4 metric features, which
contain the sepal width, the sepal length, the petal width and the petal
length of the flowers \citep{Fisher1936,Anderson1936}.

\begin{figure}[h]
  \centering
    \includegraphics[width = 0.6\textwidth]{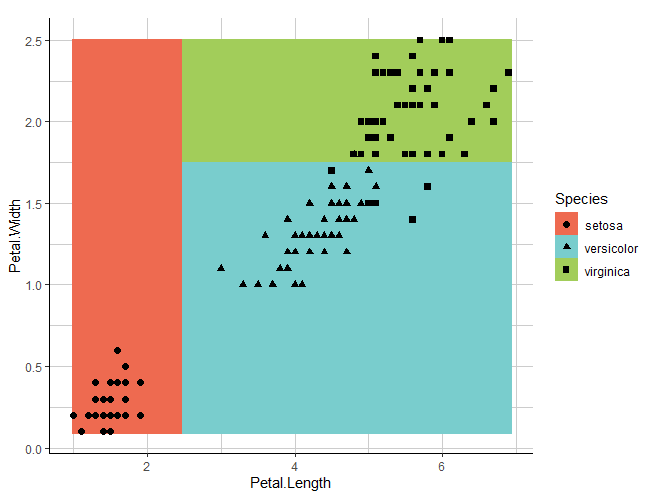}
  \caption{Iris flower data classified by a classification tree}
  \label{iris_overview}
\end{figure}

The two features petal length and petal width are known to be good
predictors for the species. A classification tree is used to predict the
three different species by these two features with the
\texttt{rpart}-package \citep{rpart}. In
Figure~\ref{iris_overview} the resulting partitioning of
the feature space is shown.

An important reason why this dataset is used here, is that the dataset
contains a lot of observations with equal feature values. For the petal
width, for example, the 150 observations have just 22 different values.
For the fitted model, quantile shift sizes of
\(q_{Petal.Width,1} = \frac{3}{151}\) and
\(q_{Petal.Width,2} = \frac{6}{151}\), respectively,\\
are chosen. Here, the \textit{shift all ties} method and the
\textit{repeatedly shift ties randomly} method are compared. For the
former method, ten repetitions where utilized to get an overall
migration matrix.

\begin{table}[h!]
\centering
\begin{tabular}{l|r r r}
\hline
  & setosa & versicolor & virginica\\
\hline
setosa & 50 & 0 & 0\\
versicolor & 0 & 48 & 6\\
virginica & 0 & 0 & 46\\
\hline
\end{tabular}
\quad
\vspace{0.2cm}
\begin{tabular}{l|r r r}
\hline
  & setosa & versicolor & virginica\\
\hline
setosa & 50 & 0 & 0\\
versicolor & 0 & 48 & 6\\
virginica & 0 & 0 & 46\\
\hline
\end{tabular}
\caption{Migration matrix for prediction changes when raising petal width by $q_{Petal.Width,1}~=~\frac{3}{151}$ (left) and $q_{Petal.Width,1}~=~\frac{6}{151}$ (right) with the \textit{shift all ties} method.}
\label{mig_mat_iris_all}
\vspace*{0.6cm}
\end{table}

\begin{table}[h!]
\centering
\begin{tabular}{l|r r r}
\hline
  & setosa & versicolor & virginica\\
\hline
setosa & 500 & 0 & 0\\
versicolor & 0 & 510 & 30\\
virginica & 0 & 0 & 460\\
\hline
\end{tabular}
\quad
\vspace{0.2cm}
\begin{tabular}{l|r r r}
\hline
  & setosa & versicolor & virginica\\
\hline
setosa & 500 & 0 & 0\\
versicolor & 0 & 480 & 60\\
virginica & 0 & 0 & 460\\
\hline
\end{tabular}
\caption{Migration matrix for prediction changes when raising petal width by $q_{Petal.Width,1}~=~\frac{3}{151}$ (left) and $q_{Petal.Width,1}~=~\frac{6}{151}$ (right) with the \textit{repeatedly shift ties randomly} method.}
\label{mig_mat_iris_random}
\end{table}

In Table~\ref{mig_mat_iris_all} it is shown that by
using the \textit{shift all ties} method, six observations change their
prediction from \textit{versicolor} to \textit{virginica}, and all other
predictions remain the same for both choices of \(q_{Petal.Width}\). In
the left migration matrix of
Table~\ref{mig_mat_iris_random} it is shown that by
using the \textit{repeatedly shift ties randomly} method, for the shift
size \(q_{Petal.Width,1}\) altogether 30 observations change their
prediction from \textit{versicolor} to \textit{virginica} in ten
repetitions, which means that just three observations change their
prediction per repetition. In the right migration matrix of
Table~\ref{mig_mat_iris_random} it is shown that for the
shift size \(q_{Petal.Width,2}\) even 60 observations change their
prediction from \textit{versicolor} to \textit{virginica} in ten
repetitions, which means that 6 observations change their prediction per
repetition, and which coincides with the results for the
\textit{shift all ties} method.

This shows that the \textit{repeatedly shift ties randomly} method is
more sensible to changes of \(q\) in general compared to the
\textit{shift all ties} method, which instead can be argued as slightly
more robust in regard to small changes of \(q\).

Furthermore, it is shown here that using the different methods changes
the resulting migration matrix, even though in this example the
difference is just in the number of observations, which change their
prediction. As argued above, the \textit{repeatedly shift ties randomly}
method treats the observations more fairly, which is why it generally
should be preferred.

\begin{figure}[h!]
\begin{subfigure}{.5\textwidth}
  \centering
  \includegraphics[width = \textwidth]{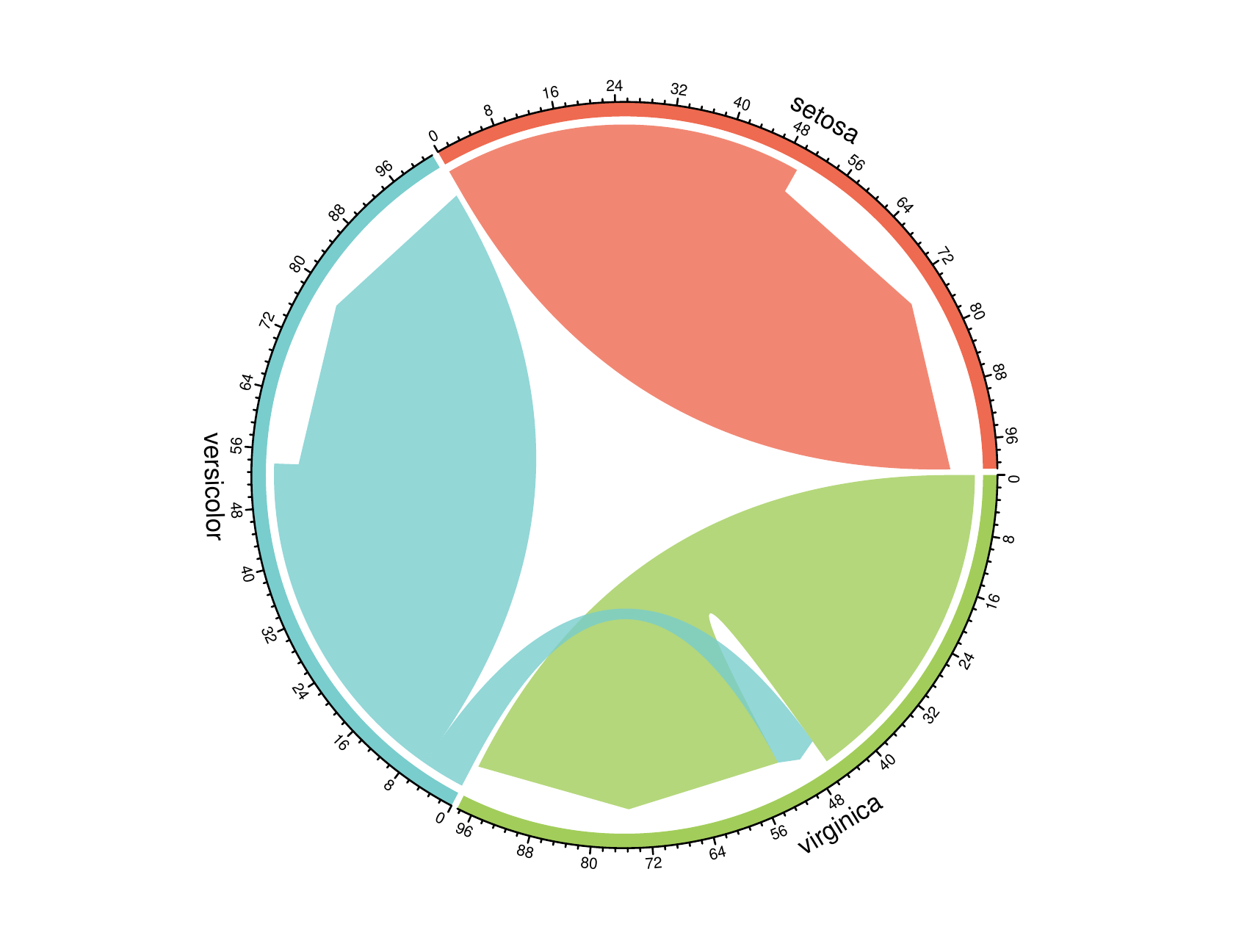}
  \caption{Chordgraph for $q_{Petal.Width,1} = 0.02$}
  \label{iris_chord_all_small}
\end{subfigure}
\hspace*{3mm}
\begin{subfigure}{.5\textwidth}
  \centering
  \includegraphics[width = \textwidth]{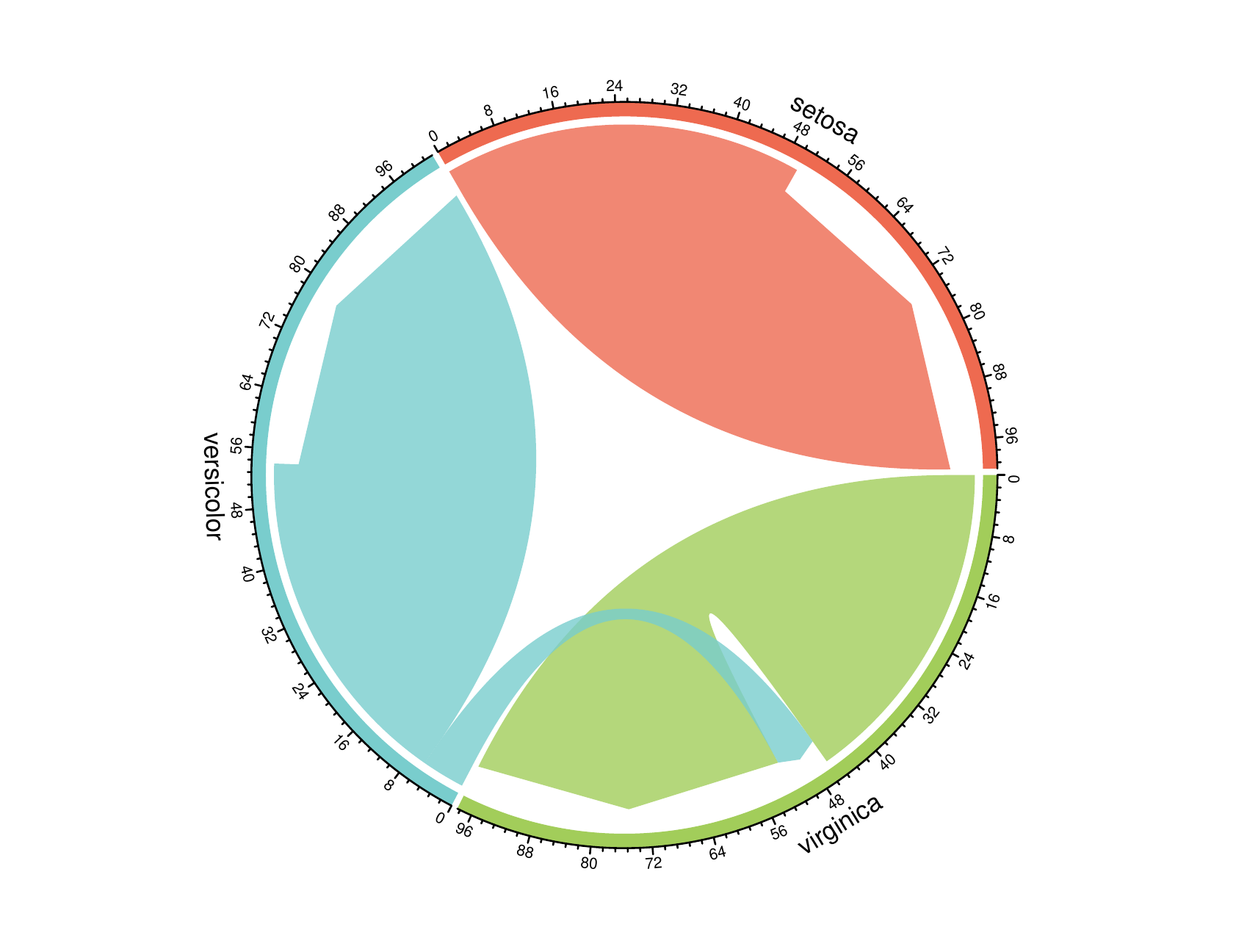}
  \caption{Chordgraph for $q_{Petal.Width,1} = 0.04$}
  \label{iris_chord_all_big}
\end{subfigure}
\caption{Chordgraphs for the \textit{shift all ties} method}
\label{iris_chord_all_fig}
\end{figure}

\begin{figure}[h!]
\begin{subfigure}{.5\textwidth}
  \centering
  \includegraphics[width = \textwidth]{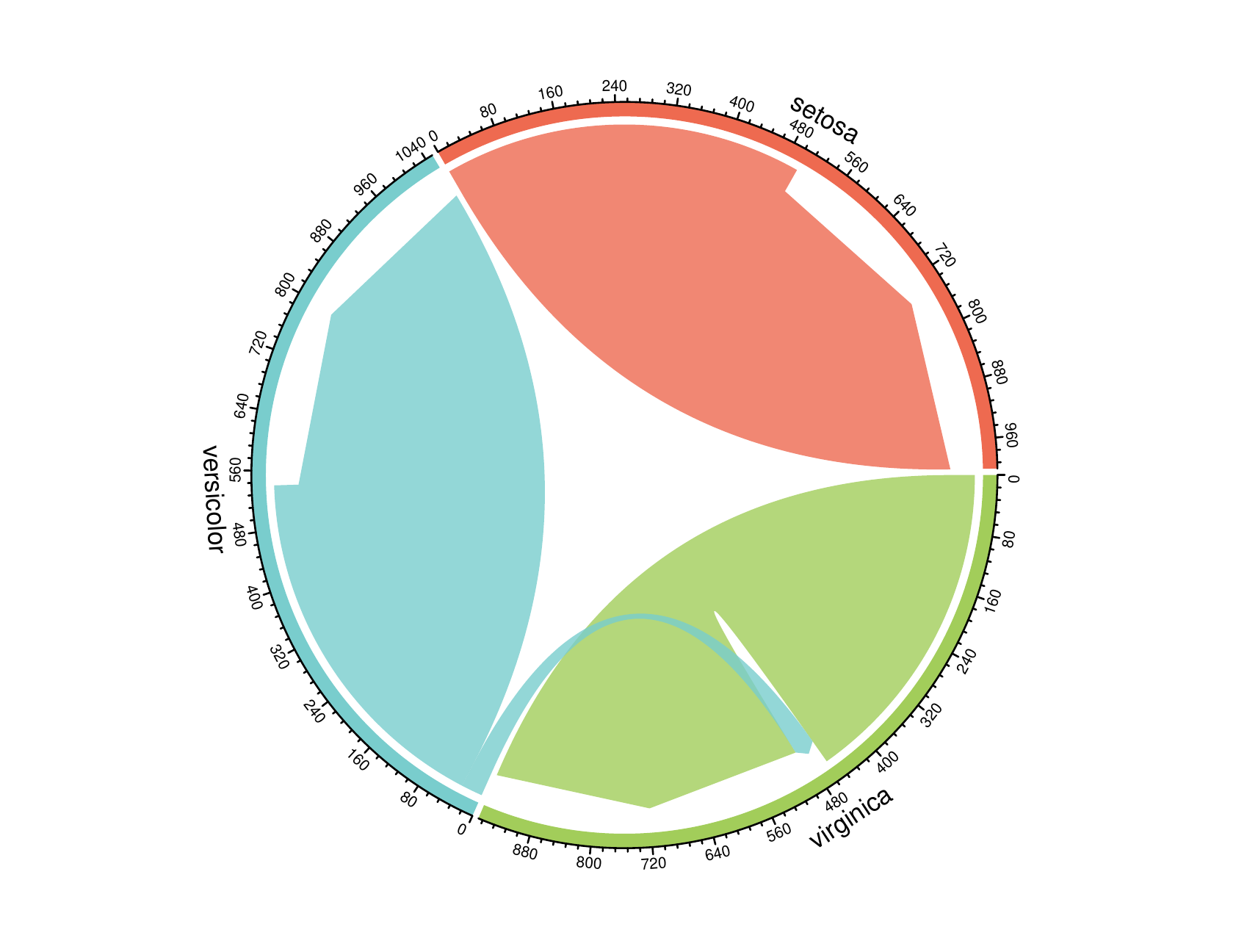}
  \caption{Chordgraph for $q_{Petal.Width,1} = 0.02$}
  \label{iris_chord_random_small}
\end{subfigure}
\hspace*{3mm}
\begin{subfigure}{.5\textwidth}
  \centering
  \includegraphics[width = \textwidth]{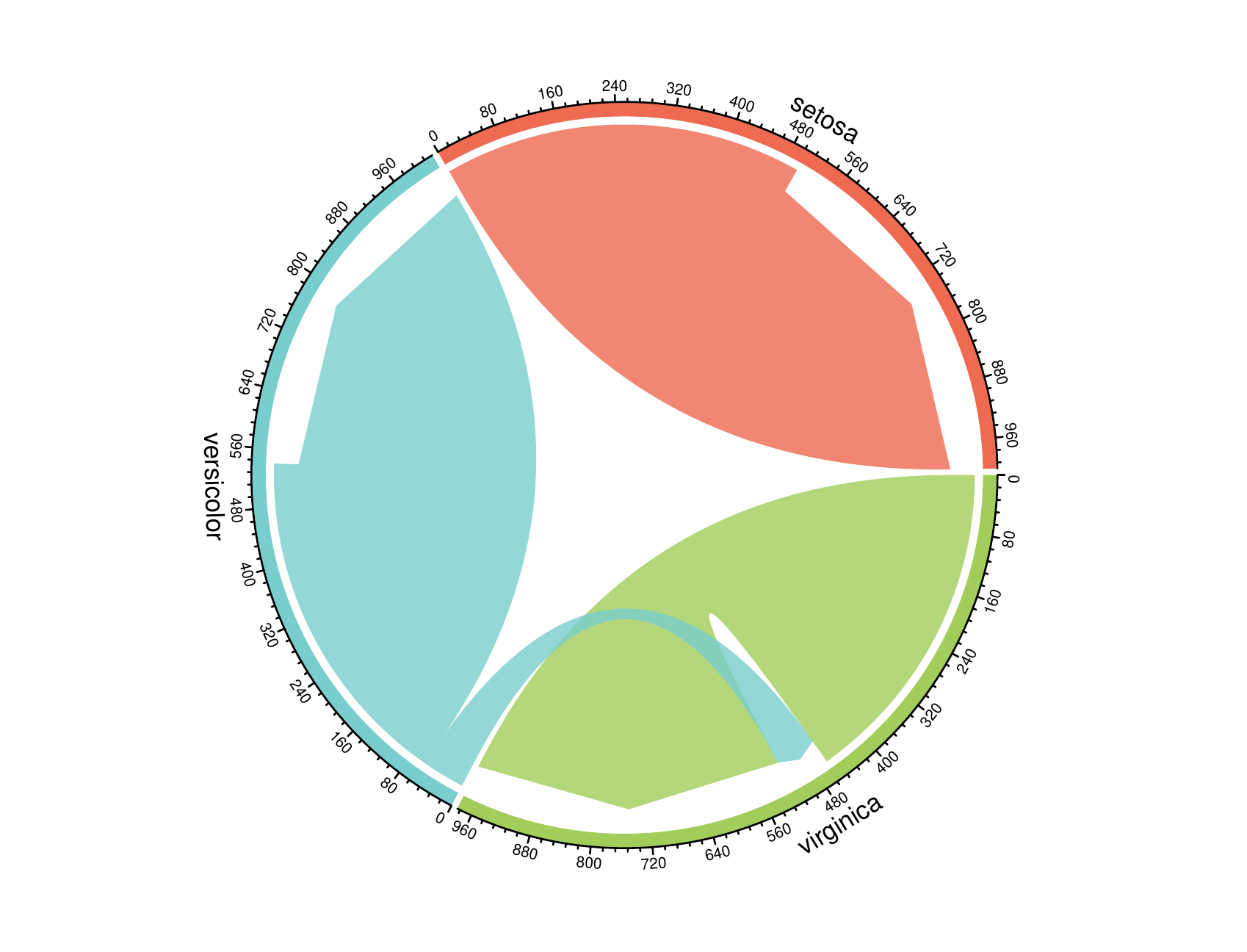}
  \caption{Chordgraph for $q_{Petal.Width,1} = 0.04$}
  \label{iris_chord_random_big}
\end{subfigure}
\caption{Chordgraphs for the \textit{repeatedly shift ties randomly} method}
\label{iris_chord_random_fig}
\end{figure}

Figure~\ref{iris_chord_all_fig} displays that the small
change in \(q_{Petal.Width}\) does actually not change the visual
impression on the found neighborhoods, as the migration matrix in the
\textit{shift all ties} method stays the same for both choices of
\(q_{Petal.Width}\).

In Figure~\ref{iris_chord_random_fig} the visual
impression (slightly) changes for the
\textit{repeatedly shift ties randomly} method, when again
\(q_{Petal.Width}\) is modified. This is also more intuitive as by
raising the absolute value of the shift size \(q\), in general the
intention is that this increases the chance that predictions do switch
and, thus, by raising the absolute value of \(q\) one would expect to
observe more changes. The main downside here is that the migration
matrix gets blown up due to the repetitions and, hence, does not
represent the amount of observations in the original dataset anymore.
This also affects the scaling of the chordgraph's axis. But as the
visual impression and interpretation are the main reasons why
chordgraphs are chosen as the visual representation method instead of
just reading the exact numbers of switched predictions - e.g., from the
migration matrix - this is rather negligible. Another downside is that
the \textit{repeatedly shift ties randomly} method takes more time to
compute than the \textit{shift all ties} method.

Overall, both methods show the same neighborhoods, and even in complex
data situations they should tend to do so, if the number of performed
repetitions is chosen appropriately. The only difference is that the
\textit{repeatedly shift ties randomly} method is more sensible to
changes in \(q\) compared to the \textit{shift all ties} method.

\hypertarget{conclusion}{%
\section{\texorpdfstring{Conclusion\label{chap:conclusion}}{Conclusion}}\label{conclusion}}

In this work, a method to determine neighborhoods between predicted
classes in a fitted model is presented, accompanied by examples
illustrating the purpose of the method and how to correctly interpret
the corresponding results. The method improves the understanding of the
partitioning of the feature space of a statistical classification model
and can be simply visualized and interpreted with the aid of
chordgraphs. The main advantage of the method surely is the gain of
insight about the partitioning of the feature space of the corresponding
classification model at hand, even though the model at hand might be a
mere black-box, hardly interpretable for practitioners.

Similar to most statistical tools, the method is an approximization of
the reality, which becomes more meaningful the better the model at hand
performs. However, the real additional value of the proposed method is
its wide and unrestricted applicability to any kind of classification
model. In general, the method can also be applied in very
high-dimensional and complex settings. The only conditions are that the
fitted model at hand builds upon a feature space and returns a
categorical output or predicted probabilities for the different response
categories.

The greatest risks when using this method are probably false
interpretations of results, as the pitfalls in this regard are manifold.
Of course, the method only describes the underlying model, and, hence,
heavily relys on its goodness-of-fit and adequacy.\\
The usage of a weak model, which badly represents reality, will almost
certainly lead to unpractical interpretations by the proposed method (as
any other model-describing methods would do as well). Another source for
potential misinterpretation could occur, if the feature manipulations
are too strong or too weak, such that some neighboring classes are
skipped or simply not found.

Nevertheless, the examples in this work show the variety of fields this
method could be applied to. First of all, QSM can help to determine
which classes are modeled close to each other by the model at hand and
can show in which order they are modeled next to each other with regard
to the manipulation. Even more, this method shows to which class
predictions generally tend to switch if one or multiple specific
feature(s) of interest are actively changed. 
Even though this does generally not allow any conclusions regarding causality and, hence, 
results have to be interpreted with caution, the method can be relevant e.g.~in clinical usage 
when a specific medication should be used to alter the features.

Since extrapolation typically is a problem in many statistical modeling
tasks and typically gets worse when the model complexity rises, care has
to be taken when large feature manipulations are used. These might force
the model to predict new classes in a region of the feature space where
data points are rather unlikely or even impossible.\\
In order to avoid this problem, it is recommended to start with rather
small feature manipulations, assuming that very small manipulations do
not create impossible observations. As it is typically hard to determine
what generally defines a ``likely'' observation, we recommend that this
should be decided manually by the user. Altogether, this manuscript aims
at providing sufficient advice to enable practitioners to safely use
this tool for meaningful interpretations.

\section{Sponsor}

This work was supported by the German Innovation Funds according to § 92a (2 )Volume V of the Social Insurance Code (§ 92a Abs. 2, SGB V - Fünftes Buch Sozialgesetzbuch), grant number: 01VSF18019. The funding body did not play any role in the design of the study and collection, analysis, and interpretation of data and in writing the manuscript.

\renewcommand\refname{References\label{references}}
\bibliography{ref.bib}

\end{document}